\documentclass[runningheads]{llncs}
\usepackage{graphicx}
\usepackage{tikz}
\usepackage{comment}
\usepackage{amsmath,amssymb} 
\usepackage{color}
\usepackage{pifont}
\usepackage{float}

\usepackage[pagebackref=true,breaklinks=true,letterpaper=true,colorlinks,bookmarks=false]{hyperref}

\usepackage{xspace}

\makeatletter
\DeclareRobustCommand\onedot{\futurelet\@let@token\@onedot}
\def\@onedot{\ifx\@let@token.\else.\null\fi\xspace}

\def\eg{\emph{e.g}\onedot}

\def\etal{\emph{et al}\onedot}
\makeatother

\graphicspath{{figure/}}
\begin{document}
\pagestyle{headings}
\mainmatter
\def\ECCVSubNumber{5697}  

\title{Efficient Video Deblurring Guided by \\Motion Magnitude} 

\titlerunning{Efficient Video Deblurring Guided by Motion Magnitude}
%
\author{Yusheng Wang\inst{1} \and
Yunfan Lu\inst{2} \and
Ye Gao\inst{4} \and 
Lin Wang\inst{2,3} \and
Zhihang Zhong\inst{1} \and
Yinqiang~Zheng\inst{1} \and
Atsushi Yamashita\inst{1}}
\authorrunning{Y. Wang et al.}
%
\institute{The University of Tokyo \\
\email{\{wang,yamashita\}@robot.t.u-tokyo.ac.jp}\\
\email{zhong@is.s.u-tokyo.ac.jp}, \email{yqzheng@ai.u-tokyo.ac.jp}
\and
AI Thrust, Information Hub, HKUST Guangzhou \and Department of Computer Science and Engineering, HKUST \\
\email{\{yunfanlu,linwang\}@ust.hk}
\and
Tokyo Research Center, Huawei \\
\email{gaoye1984@yahoo.co.jp}
}

\maketitle

\begin{abstract}
 Video deblurring is a highly under-constrained problem due to the spatially and temporally varying blur. An intuitive approach for video deblurring includes two steps: a) detecting the blurry region in the current frame; b) utilizing the information from clear regions in adjacent frames for current frame deblurring. 
 To realize this process, our idea is to detect the pixel-wise blur level of each frame and combine it with video deblurring. To this end, we propose a novel framework that utilizes the motion magnitude prior (MMP) as guidance for efficient deep video deblurring. Specifically, as the pixel movement along its trajectory during the exposure time is positively correlated to the level of motion blur, we first use the average magnitude of optical flow from the high-frequency sharp frames to generate the synthetic blurry frames and their corresponding pixel-wise motion magnitude maps. 
 We then build a dataset including the blurry frame and MMP pairs. The MMP is then learned by a compact CNN by regression. The MMP consists of both spatial and temporal blur level information, which can be further integrated into an efficient recurrent neural network (RNN) for video deblurring.  We conduct intensive experiments to validate the effectiveness of the proposed methods on the public datasets.  Our codes are available at \url{https://github.com/sollynoay/MMP-RNN}. 
\end{abstract}

\section{Introduction}

Video deblurring is a classical yet challenging problem that aims to restore consecutive frames from the spatially and temporally varying blur. The problem is highly ill-posed because of the presence of camera shakes, object motions, and depth variations during the exposure interval. 
Recent methods using deep convolutional networks (CNNs) have shown a significant improvement in video deblurring performance. Among them, alignment-based methods usually align the adjacent frames explicitly to the current frame and reconstruct a sharp frame simultaneously or by multiple stages~\cite{Su2017,Wang2019,Zhou2019,Pan2020,Li2021}. Such methods show stable performance for video deblurring; however, the alignment or warping processes require accurate flow estimation, which is difficult for blurry frames, and usually the computational cost is relatively high. Recurrent neural network (RNN)-based methods achieve deblurring by passing information between adjacent frames~\cite{Kim2017,Nah2019,Zhong2020}; such methods usually have lower computation costs but with lower restoration quality compared with the alignment-based methods. 

So far, many deblurring methods have utilized priors to improve the restoration quality. The priors usually come from optical flows or image segmentation. Inter-frame optical flow-based methods~\cite{Pan2020,Li2021} directly borrow pixel information from the adjacent frames to improve the quality of the current frame. By contrast, intra-frame optical flow estimates the pixel movement during the exposure time of a blurry frame.  Intra-frame optical flow-based methods~\cite{Gong2017,Argaw2021} estimate the flow during the exposure time, and usually restore the sharp frame by energy minimization optimization. Image segmentation-based methods~\cite{Cho2007,Bar2007,Wulff2014,Ren2017,Shen2019} utilize motion, semantic or background and foreground information to estimate the prior for deblurring.  

However, previous priors for video deblurring suffer from one or more of the following problems.
First, image segmentation and inter-frame optical flow estimation on blurry images are error-prone. And it is necessary to estimate the prior and implement video deblurring simultaneously via complex energy functions \cite{Portz2012,Kim2015,Ren2017}.
Second, inter-frame optical flow estimation requires heavy computational cost. For instance, the state-of-the-art (SoTA) methods, PWC-Net requires 181.68 GAMCs \cite{Sun2018PWC-Net} and RAFT small model (20 iterations) requires 727.99 GMACs \cite{Teed2020} on the 720P (1280 x 720) frames. However, a typical efficient video deblurring method is supposed to have 50$\sim$300 GMACs. 
Third, motion blur is directly correlated to the intra-frame optical flow. Although it is possible to directly estimate the intra-frame optical flow from a single image \cite{Gong2017,Argaw2021}, the restored results suffer from artifacts.

An intuitive way for video deblurring is to detect the blurry region in the image first and then utilize information of clear pixels from adjacent frames. The detection of blurry region task can be considered as a blur estimation problem, which separates the image into binarized sharp and blurry regions, or directly tells the blur level of each pixel. Previous blur estimation methods, \eg,~\cite{Devy2013,Shi2014}, usually binarize the image into sharp and blurry regions and manually label the regions as it is difficult to determine the blur level of each pixel automatically. However, the labelling process may be inaccurate and usually requires much human effort. 

In this work, we propose a motion magnitude prior (MMP) to represent the blur level of a pixel which can determine the pixel-wise blur level without manually labelling. Recent works use high frequency sharp frames to generate a synthetic blurry frame \cite{Nah2017}. Inspired by this, the pixel movement during the exposure time of a blurry frame, i.e., the MMP, can be estimated from the average magnitude of bi-directional optical flow from the high-frequency sharp frames. The value of MMP positively correlated to the level of motion blur. We propose a compact CNN to learn the MMP.

\begin{figure}[tb]
	\centering
	
	{\includegraphics[width=0.9\columnwidth]{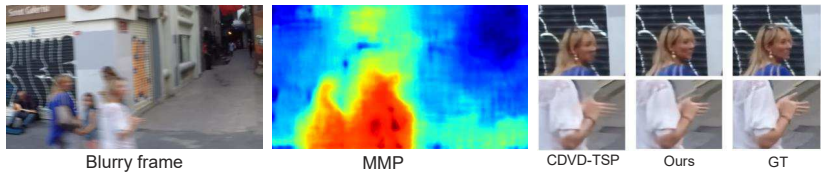}}
	\caption{The blurry frame and the estimated MMP with results compared to the SoTA method \cite{Pan2020}. }
	\label{fig:catch}
\end{figure}

The proposed MMP can directly indicate the \textit{spatial distribution} of blur level in one image. Besides, if the overall value of MMP is low, the image is sharp, and vice versa. \textit{Temporal information} is also included in the prior. The CNN can be further merged into a  \textit{spatio-temporal network} (RNN) for video deblurring. For convenience, we use an efficient RNN with residual dense blocks (RDBs) \cite{Zhang2018,Zhang2020} as our backbone video deblurring network (Sec.~\ref{sec:rnn}). For the utilization of MMP, we design three components: a) for the intra-frame utilization, we propose a motion magnitude attentive module (MMAM) to inject the MMP into the network; b) for the inter-frame utilization, different from other RNN-based methods \cite{Kim2017,Nah2019,Zhong2020} which only pass the deblurred features to the next frame, we also pass the features before deblurring to the next frame. The blurry frame with pixels of different blur level is weighted by the MMAM, as such, pixels under low blur level can be directly utilized by the next frame; c) for loss-level utilization, since the motion magnitude of network output can reflect the deblur performance. That is, the sharper the image is, the lower the average score is. Therefore, we can also use the prior as a loss term to further improve the optimization. Figure~\ref{fig:catch} is an example of the learned MMP from the network and the estimation result. High quality results can be generated from the proposed framework.

In summary, our contributions are three folds. (I)  We propose a novel motion magnitude prior for blurry image and its lightweight generation method. To the best of our knowledge, this is the first work to apply motion magnitude prior for the video deblurring task. (II) We propose a motion magnitude prior guided network, which utilizes the prior at intra-frame level, inter-frame level and loss level. (III) Our proposed method achieves competitive results with SoTA methods while with relatively lower computational cost.

\section{Related Works}

\subsection{Prior for Deblurring}
As mentioned above, image segmentation, inter-frame optical flow and intra-frame optical flow have been frequently used as priors in the deblurring tasks.

\noindent  \textbf{Image Segmentation} 
In early studies, segmentation priors have been proposed for dynamic scene deblur. Cho \etal~\cite{Cho2007} segmented the images into multiple regions of homogeneous motions while Bar \etal~\cite{Bar2007} segmented the images into foreground and background layers. Based on~\cite{Bar2007} , Wulff \etal~ \cite{Wulff2014} focused on estimating the parameters for both foreground and background motions. Using segmentation priors can handle the dynamic scenes; however, with simple models, it is difficult to segment the non-parametrically varying complex motions. Ren \etal~\cite{Ren2017} exploited semantic segmentation which significantly improves the optical flow estimation for the deblurring tasks. Shen \etal~\cite{Shen2019} proposed a human-aware deblurring method by combining human segmentation and learning-based deblurring. Although these methods on the basis of physical models show promising results, the deblurring performance is highly related to the blur kernel estimation results. That is, inaccurate estimation of blur kernels results in severe artifacts. 

\noindent  \textbf{Inter-frame optical flow} 
The inter-frame optical flows are usually directly estimated on blurry frames, and are used to warp the adjacent frames to the center frame before inputting to the network \cite{Su2017,Pan2020,Li2021}. The flow from blurry frame is inaccurate. To solve the problem, Pan \etal~\cite{Pan2020} proposed a method to estimate optical flow from the intermediate latent frames using a deep CNN and restore the latent frames based on the optical flow estimations. A temporal sharpness prior is applied to constrain the CNN model to help the latent frame restoration. However, calculating inter-frame optical flow requires heavy computation and the temporal sharpness prior cannot deal with all frames with no sharp pixels. 

\noindent  \textbf{Intra-frame optical flow} 
With deep learning, it is even possible to directly estimate intra-frame optical flows from a blurry frame \cite{Sun2015,Gong2017,Argaw2021}, followed by energy function optimizations to estimate the blur kernels. However, the optimizations are usually difficult to solve and require a huge computational cost. The restored images also suffer from artifacts. Moreover, the definition of intra-frame is ambiguous as, during the exposure time, the movement of one pixel may be non-linear, which cannot be represented by a 2D vector. 

\noindent  \textbf{Others} 
Statistical priors and extreme channel priors have shown their feasibility in single image deblurring \cite{Shan2008,Xu2013,Pan2016,Yan2017}. Such methods are valid under certain circumstances, but are sensitive to noise and still require accurate estimation of the blur kernels.
Differently, we propose a motion magnitude prior learned by a compact network. The prior shows pixel-wise blur level of the blurry image and can be easily applied to the video deblurring framework. It can detect which part of the image is blurry and to what extent it is blurred. 

\subsection{Blur Estimation}
Blur estimation has been studied for the non-uniform blur. It is similar to image segmentation prior by specifically segmenting the image into blur and non-blur regions. In \cite{Devy2013}, horizontal motion or defocus blurs were estimated and the image is deconvolved for deblurring. Shi \etal~\cite{Shi2014} studied the effective local blur features to differentiate between blur and sharp regions which can be applied to the blur region segmentation, deblurring and blur magnification. The images are usually separated into blur and non-sharp regions; however, it is difficult to separate an image under a binarized way for dynamic scenes. 
Instead of classifying the pixels into blurry and sharp regions, we use a continuous way to represent the blur maps by estimating the motion magnitude of each pixel. 

\subsection{DNN-based Deblurring}
For single image deblurring, SoTA methods apply self-recurrent module on the multi-scale, multi-patch, or multi-stage to solve the problem \cite{Nah2017,Tao2018,Gao2019,Suin2020,Zamir2021}. Despite the high performance, they usually require a large computational cost. For video deblurring, the problem is less ill-posed and can be solved with less computational cost. The learning-based video deblurring methods can be grouped into alignment-based methods \cite{Su2017,Wang2019,Zhou2019,Pan2020,Li2021} and RNN-based methods \cite{Kim2017,Nah2019,Zhong2020}. The former usually explicitly aligns the adjacent frames to the center frame for deblurring. 
This can be achieved by directly warping the adjacent frames to the center frame \cite{Su2017}, warping the intermediate latent frames, or both \cite{Pan2020,Li2021}. 
Wang \etal~\cite{Wang2019} implemented the deformable CNN \cite{Dai2017} to realize the alignment process. Zhou \etal~\cite{Zhou2019} proposed a spatio-temporal filter adaptive network (STFAN) for alignment and deblurring in a single network. Son et al. blur-invariant motion estimation
learning to avoid the inaccuracy caused by optical flow estimation on blurry frames \cite{Son2021}. 
Although the alignment-based methods can achieve relatively high performance, the alignment process is usually computationally inefficient. In addition, the alignment has to be accurate, otherwise it may degrade the performance of deblurring. To ensure high quality results, multi-stage strategies were applied by iteratively implementing the alignment-deblurring process, which may lead to huge computational cost \cite{Pan2020,Li2021}. 

On the other hand, RNN-based methods transfer information between the RNN cells and usually have lower computational cost. Kim \etal~\cite{Kim2017} propose an RNN for video deblurring by dynamically blending features from previous frames. Nah \etal~\cite{Nah2017} iteratively updated the hidden state with one RNN cell before the final output. Zhong \etal~ \cite{Zhong2020} proposed an RNN network using RDB backbone with globally fusion of high-level features from adjacent frames and applied attention mechanism to efficiently utilize inter-frame information. 
In this paper, we propose an RNN with RDBs by utilizing the MMP. We use a compact CNN to estimate the MMP to reduce the computational cost. The MMP consists of spatial and temporal information, which can improve the effectiveness of information delivery between RNN cells. Our proposed method outperforms the SoTA performance.

\section{The Proposed Approach}

\subsection{Motion Magnitude Prior}\label{sec:mmp}

\begin{figure}[tb]
	\centering
	
	{\includegraphics[width=1.0\columnwidth]{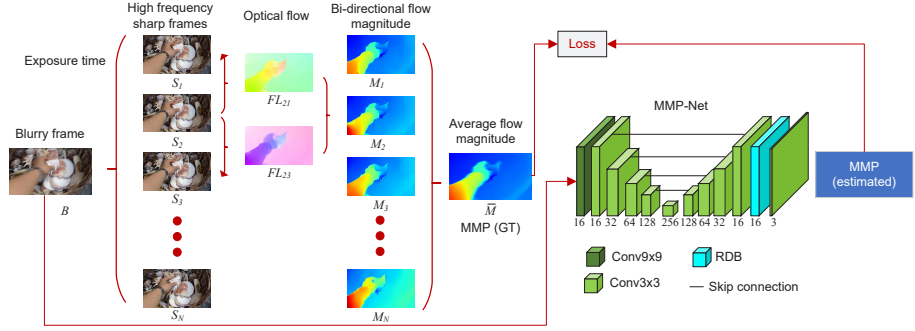}}
	\caption{Motion magnitude prior. High frequency sharp frames are used to generate synthetic blurry frames. Meanwhile, we estimate the bi-directional optical flows for each frame and calculate the average magnitude of bi-directional optical flows. Then, we take the average of the motion magnitude maps of all the latent frames. Finally, we use a modified UNet-like structure to learn the motion magnitude prior by regression. }
	\label{fig:prior}
\end{figure}

\noindent  \textbf{Motion Magnitude Prior from Optical Flow} 
In this section, we introduce preparing the ground truth of motion magnitude prior (MMP). 
For the learning-based deblurring methods, the datasets for training and validation are usually synthesized by high-frequency sharp frames \cite{Su2017,Nah2017,Wang2019}. It is based on the fact that images tend to be blurry after accumulating multiple sharp images with slight misalignment caused by motion~\cite{Hirsch2011}. 
The blur level of each pixel in the blurry frame is positively correlated to the motion magnitude of the pixel during exposure time. Although directly measuring the motion magnitude of one pixel is difficult, it is possible to be calculated by latent sharp frames during exposure time. 
Inspired by this, we generate blurry frame and blur level map pairs based on high-frequency sharp image sequence, as shown in Fig.~\ref{fig:prior}.
Denoting the blurry image as $B$ and sampled sharp image as \{$S_1,S_2,\dots,S_N$\}, the blurry image simulation process can be represented as follows.
\begin{equation}
	B = c \left( \frac{1}{N}\sum_{i=1}^{N}S_i \right),
\end{equation}
where $c(.)$ refers to the camera response function (CRF) \cite{Tai2013}. During exposure time, we sample $N$ sharp images to generate a blurry image. 
To measure the movement of pixels, we calculate the optical flow between the sharp frames. We use bi-directional flows to represent the pixel movement of one sharp frame. For instance, as shown in Fig.~\ref{fig:prior}, for frame 1, we calculate the optical flow $FL_{21}$ and $FL_{23}$ to represent the pixel movement condition. For each pixel $(m,n)$, the optical flow between frame $i$ and $j$ in $x$ and $y$ direction can be represented as $u_{i,j}(m,n)$ and $v_{i,j}(m,n)$, respectively. We denote the motion magnitude $M$ for frame $i$ as follows.

\begin{equation}\label{eq:2}
	M_{i}(m,n)=\frac{\sqrt{u_{i,i-1}^2(m,n)+v_{i,i-1}^2(m,n)}+\sqrt{u_{i,i+1}^2(m,n)+v_{i,i+1}^2(m,n)}}{2}.
\end{equation}

For frame 1 and frame $N$, we only use $FL_{12}$ and $FL_{N,N-1}$ for calculation, respectively. Then, to acquire the pixel-wise motion magnitude for the synthetic blurry frame, we calculate the average movement during exposure time.
\begin{equation}
	\bar{M} = \frac{1}{KN}\sum_{i=1}^{N}M_{i},
\end{equation}
where $\bar{M}$ refers to the motion magnitude or blur level map for the blurry frame. $K$ is used to normalize the value at each position in MMP to 0$\sim$1. $K$ is determined by the maximum value of the MMP before normalization in the training dataset which was set to 15. 


\noindent \textbf{The Learning of Motion Magnitude Prior} 
In this paper, we use the GOPRO raw dataset \cite{Nah2017} to generate blurry image and MMP pairs. The GOPRO benchmark dataset used 7$\sim$13 successive sharp frames in raw dataset to generate one blurry frame. We also use 7$\sim$13 to generate the blurry image and MMP pairs. We use the SoTA optical flow method RAFT \cite{Teed2020} to estimate the optical flow. 
We propose a compact network to learn the blur level map by regression. We apply a modified UNet \cite{Ronneberger2015}, as shown in Fig.~\ref{fig:prior}. 
At the beginning, we use a 9 $\times$ 9 convolution layer to enlarge the reception field. The features are downsampled and reconstructed with a UNet-like structure. Then, a residual dense block (RDB) is used to refine the result. 
To process a 720P (1280 $\times$ 720) frame, the computational cost is only 38.81 GMACs with a model size of 0.85M parameters. At last, we can estimate the MMP for each frame using the compact network. In this paper, we apply the MMP as guidance to the video deblurring network. We will describe it in the following section.

\subsection{Motion Magnitude Prior-Guided Deblurring Network}\label{sec:rnn}

In this section, we describe using MMP as guidance to video deblurring network. 

\begin{figure}[tb]
	\centering
	\includegraphics[width=1.0\columnwidth]{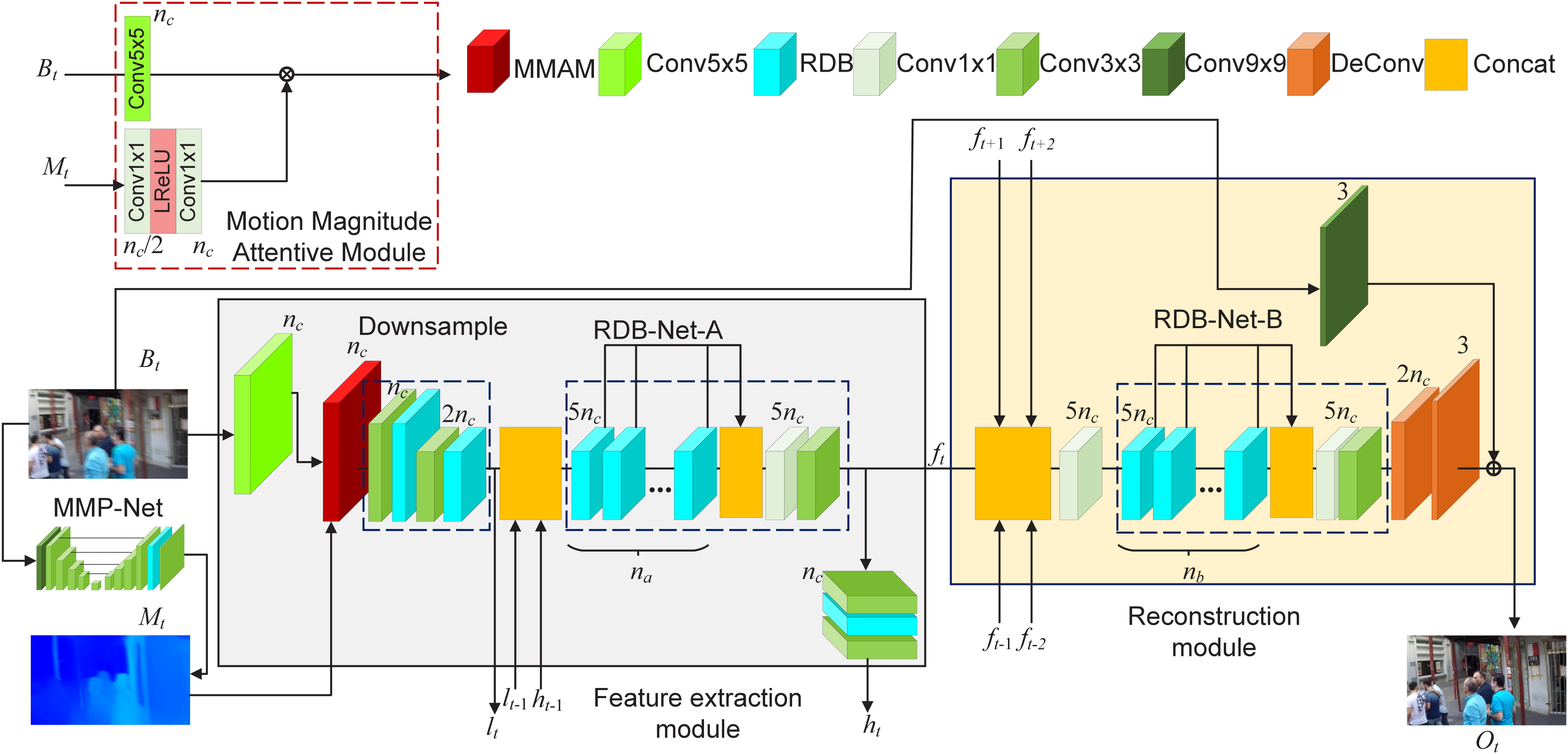}
	\caption{The structure of MMP-RNN. Both center frame $I_t$ and corresponding blur prior $B_t$ estimated by MMP-Net are passed into RNN cell. We transmit both non-deblurred feature $l$ and deblurred feature $h$ from the previous frame to the next frame. Deblurred features $f$ from the current frame and the adjacent frames are fused globally in decoder to generate final output image. }
	\label{fig:rnn}
\end{figure}

\noindent\textbf{Network Structure}\label{sec:overview} The structure of MMP-RNN is shown in Fig.~\ref{fig:rnn}. $n_c$ refers to the channel dimension. The center frame $B_t$ is first inputted into MMP-Net to estimate the MMP $M_t$. Then, $B_t$ and $M_t$ are passed to RNN cell to extract features. Non-deblurred features $l_{t-1}$ and deblurred features $h_{t-1}$ from the previous frame are delivered to feature extraction module (FEM). In this paper, we also use the deblurred features from adjacent frames globally to reconstruct output images. Features $f_{t-2}$, $f_{t-1}$ from the past frames and $f_{t+1}$, $f_{t+2}$ from the future frames with the current frame features $f_t$ are inputted to reconstruction module (RM) for image reconstruction. We denote the output result as $O_t$. 
RDBs have high performances in low-level tasks which efficiently preserve features and save the computational cost \cite{Zhang2018,Zhang2020,Zhong2020}. In this work, we use RDBs as the backbone for downsampling, feature extraction and implicit alignment. 

The FEM receives center frame $B_t$ and corresponding MMP $M_t$ with features $l_{t-1}$ and $h_{t-1}$ from the past frames to extract features of current frame. The structure of FEM is shown in Fig.~\ref{fig:rnn}. The Motion Magnitude Attentive Module (MMAM) is used to fuse the information of $B_t$ and $M_t$. Here, we pass the non-deblurred features $l_t$ to the next FEM and receive $l_{t-1}$ from the previous FEM. The non-deblurred features and deblurred features are then concatenated and passed through RDB-Net-A with $n_a$ RDBs. At last, deblurred features $h_t$ are passed to the next FEM and $f_t$ are used for image reconstruction. 

RM is designed to globally aggregate high-level features for image reconstruction. We concatenate the features \{$f_{t-2},\dots,f_{t+2}$\} and squeeze them in channel direction using 1 $\times$ 1 convolution operation. We use $n_b$ RDBs in RDB-Net-B to implicitly align the features from different frames and then apply convolution transpose operation to reconstruct the image. A global skip connection is added to directly pass $I_t$ to the output after 9 $\times$ 9 convolution operation. This can better preserve the information of the center frame.    




\noindent \textbf{Motion Magnitude Attentive Module}\label{sec:sft} 
Both the blurry and sharp pixels are important to our task. The blurry pixels are the region to be concentrated for deblurring and the sharp pixels can be utilized for deblurring of adjacent frames. Especially for sharp pixels, the value of MMP is close to 0, directly multiplying MMP to image features may lose information of sharp features. To better utilize MMP, we use MMAM to integrate it with image features. The structure of MMAM layer is shown in Fig.~\ref{fig:rnn}. The MMP is passed through two 1 $\times$ 1 convolution layers to optimize MMP and adjust the dimension.
The MMP is transformed to tensor $\gamma$ . 
The operation to integrate $\gamma$ and the feature from $B$, $x$ can be represented as $
x^{out} = \gamma\otimes x^{in}$,
where $\otimes$ refers to element-wise multiplication. The whole operation can better integrate the MMP information with the blurry image. 

\noindent \textbf{Feature Transmission}\label{sec:trans}
We transmit two kinds of features $l$ and $h$ between FEM of the adjacent frames. The non-deblurred feature $l_t$ consists of the information only from the center frame before integration. By contrast, $h_t$ possesses features from the previous frame. The features are integrated in the network as follows. 
\begin{equation}
	a_t=CAT(l_t,l_{t-1},h_{t-1}),
\end{equation}
where $CAT$ refers to concatenation operation. The RNN-based methods implicitly fusing information from previous frames which may sometimes cause lower performance compared to alignment-based methods. Passing non-deblurred features only consisting of the current frame with its blur level information can improve the overall performance of the network. 

\noindent\textbf{Motion Magnitude Loss}\label{sec:loss-functions} 
The proposed MMP-Net can estimate pixel-wise motion magnitude of the image. If $O_t$ is an ideal sharp image, inputting $O_t$ into MMP-Net should get a prior with all zeros. In this work, we propose a loss function based on the idea as follows. 
\begin{equation}
	\mathcal{L}_{MM} = \frac{1}{mn}\sum_{i=0}^{m-1}\sum_{j=0}^{n-1}M_{i,j}(O_t),
\end{equation}
 where $M(O_t)$ refers to the MMP of the output image $O_t$. Theoretically, by minimizing the average motion magnitude of $O_t$, it can generate ideal sharp image.
 
 We also consider two kinds of content loss functions for training. We use a modified Charbonnier loss ($\mathcal{L}_{char}$) \cite{Charbonnier1994} and gradient loss ($\mathcal{L}_{grad}$) as content loss.
 \begin{equation}
 	\mathcal{L}_{char} = \frac{1}{mn}\sum_{i=0}^{m-1}\sum_{j=0}^{n-1}
 	\sqrt{\sum_{ch}^{r,g,b}(I_{i,j,ch}-O_{i,j,ch})+\epsilon^2},
 \end{equation}
 where $\epsilon$ is a small value which is set to 0.001. $m$ and $n$ are the width and height of the image. $I$ refers to the ground truth and $O$ is the estimated image.
 \begin{equation}
 	\mathcal{L}_{grad} = \frac{1}{mn}\sum_{i=0}^{m-1}\sum_{j=0}^{n-1}(G_{i,j}-\hat{G}_{i,j})^2,
 \end{equation}
 where $G$ and $\hat{G}$ refers to the image gradient of $I$ and $O$, respectively. Then, the total loss is written as follows.
 \begin{equation}
 	\mathcal{L}=\mathcal{L}_{char}+\lambda_1 \mathcal{L}_{grad}+\lambda_2 \mathcal{L}_{MM},
 \end{equation}
 where the weight $\lambda_1$ and $\lambda_2$ are set to 0.5 and 1.0 in our experiments.

\section{Experiment}

\subsection{MMP-Net}\label{sec:deep-image-prior}

\noindent  \textbf{Dataset Generation} 
To learn the deep image prior, we utilize GOPRO dataset \cite{Nah2017}. The raw GOPRO dataset consists of 33 high-frequency video sequences with 34,874 images in total. In GOPRO benchmark dataset, the sharp images are used to generate synthetic dataset with 22 sequences for training and 11 sequences for evaluation with 2,103 training samples and 1,111 evaluation samples. We use the same data separation as GOPRO benchmark dataset to build the MMP dataset to avoid information leakage during video deblurring. We use 7$\sim$11 consecutive sharp frames to generate one blurry frame and corresponding MMP. We generate 22,499 training samples (22 sequences) and use the original GOPRO test dataset. We trim the training dataset by 50\% during training. 

\noindent  \textbf{Implementation details} 
We train the model for 400 epochs with a mini-batch of size 8 using ADAM optimizer \cite{Kingma2015} with initial learning rate 0.0003. The learning rate decades by half after 200 epochs. The patch size is set to $512 \times 512$ for training and validation. 
The loss $\mathcal{L}_{1} = \frac{1}{mn}\sum_{i=0}^{m-1}\sum_{j=0}^{n-1}|M_{i,j}-\hat{M}_{i,j}|$
 is used for MMP training and as a metric for test. Here, $M_{i,j}$ refers to the value of estimated MMP at position $i,j$ and $\hat{M}$ refers to the GT.

\noindent  \textbf{Results} 
\begin{figure}[t!]
	\centering
	\includegraphics[width=.75\columnwidth]{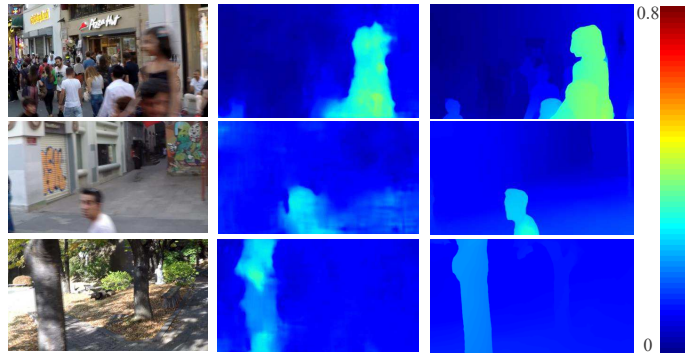}  
	\caption{Examples of motion magnitude estimation on GOPRO test dataset. Each column refers to the input blurry frames, the estimated results, and GT, respectively.}
	\label{fig:gopro_blm}
\end{figure}
\begin{figure}[tb]
	\centering
	
	{\includegraphics[width=.75\columnwidth]{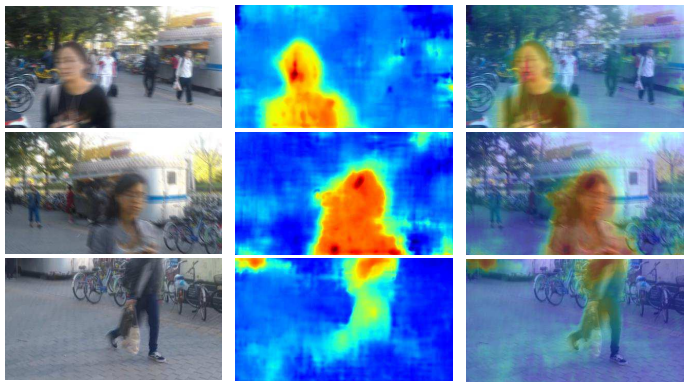}}
	\caption{Examples of test results on HIDE dataset \cite{Shen2019}. The model was trained on GOPRO dataset and tested on HIDE dataset. Each column refer to the input blurry images, estimated results, and the estimated results overlaid to the input images. It indicates that the proposed method can successfully detect salient blurry region on other datasets.}
	\label{fig:hide}
\end{figure}
\setlength{\tabcolsep}{12pt}
\begin{table}[t!]
\begin{center}
\caption{The training result of MMP-Net.}

\label{table:mmp}
\begin{tabular}{lccc}
				\hline
				 & Train & Test (512 patch) & Test (Full image)  \\
				\hline
				$\mathcal{L}_1$  & 0.0137 & 0.0169 & 0.0192  \\
				\hline
			\end{tabular}
\end{center}

\end{table}
\setlength{\tabcolsep}{1.4pt}
The training results are listed in Table~\ref{table:mmp}. We visualize our results on GOPRO test dataset in Fig.~\ref{fig:gopro_blm}. To prove the generality of the proposed method, we also test the GOPRO trained model to HIDE dataset~\cite{Shen2019}. HIDE dataset contains blurry images of human. As shown in Fig.~\ref{fig:hide}, our proposed method can also detect the blur caused by human motion on other dataset.

\subsection{Video Deblurring}\label{sec:video-deblurring}
\setlength{\tabcolsep}{1.4pt}
\noindent  \textbf{Datasets} 
We test the proposed video deblurring method on two public datasets, GORPRO benchmark dataset \cite{Nah2017} and beam-splitter datasets (BSD) \cite{Zhong2020}. BSD is a dataset of images from real world by controlling the length of exposure time and strength of exposure intensity during video shooting using beam-splitter system \cite{Jiang2019}. It can better evaluate deblurring performance in real scenarios. We use the 2ms-16mes BSD that the exposure time for sharp and blurry frames are 2~ms and 16~ms, respectively. The training and validation sets have 60 and 20 video sequences with 100 frames respectively, and the test set has 20 video sequences with 150 frames. The size of the frames is 640 $\times$ 480.

\noindent  \textbf{Implementation details} 
We train the model using ADAM optimizer with a learning rate of 0.0005. We adopt cosine annealing schedule \cite{Loshchilov2017} to adjust learning rate during training. We sample 10-frame 256 $\times$ 256 RGB patch sequences from the dataset to construct a mini-batch of size 8 with random vertical and horizontal flips as well as $90^\circ$ rotation for data augmentation for training. We train 1,000 epochs for GOPRO and 500 epochs for BSD, respectively. 
It is worth mentioning that we try to train the other methods using publicly available codes by ourselves and keep the same hyper-parameters if possible. For CDVD-TSP, we use the available test images for GOPRO and keep the training strategy used in \cite{Pan2020} for BSD.

\setlength{\tabcolsep}{6pt}
\begin{table}[t!]
\begin{center}
\caption{Quantitative results on GOPRO.}
\label{table:gopro}
\scalebox{1.0}{
\begin{tabular}{lccccc}
				\hline
				Model & PSNR & SSIM & GMACs & Param & Time (s) \\
				\hline
				SRN \cite{Tao2018} & 29.94 & 0.8953  & 1527.01 & 10.25 & 0.173\\
				DBN \cite{Su2017} & 28.55 & 0.8595 & 784.75 & 15.31 & 0.128\\
				IFIRNN ($c2h3$) \cite{Nah2019}  & 29.69 & 0.8867 & 217.89 & 1.64 & 0.034\\
				ESTRNN ($C_{70}B_7$) \cite{Zhong2020}  & 29.93 & 0.8903 & 115.19& 1.17 & 0.021\\
				ESTRNN ($C_{90}B_{10}$) \cite{Zhang2020} & 31.02 & 0.9109 & 215.26& 2.38 & 0.035\\
				CDVD-TSP \cite{Pan2020}  & 31.67 & 0.9279 & 5122.25 & 16.19 & 0.729\\
				\hline \hline
				MMP-RNN ($A_{3}B_{4}C_{16}F_{5}$) & 30.48 & 0.9021 & 136.42& 1.97 & 0.032\\
				MMP-RNN ($A_{7}B_{8}C_{16}F_{8}$) & 31.71 & 0.9225 & 204.19 & 3.05 & 0.045\\
				MMP-RNN ($A_{9}B_{10}C_{18}F_{8}$) & \textbf{32.64} & \textbf{0.9359} & 264.52 & 4.05 & 0.059\\
				\hline
			\end{tabular}
			}
\end{center}
\end{table}

\begin{figure}[tb]
	\centering
	
	{\includegraphics[width=0.85\columnwidth]{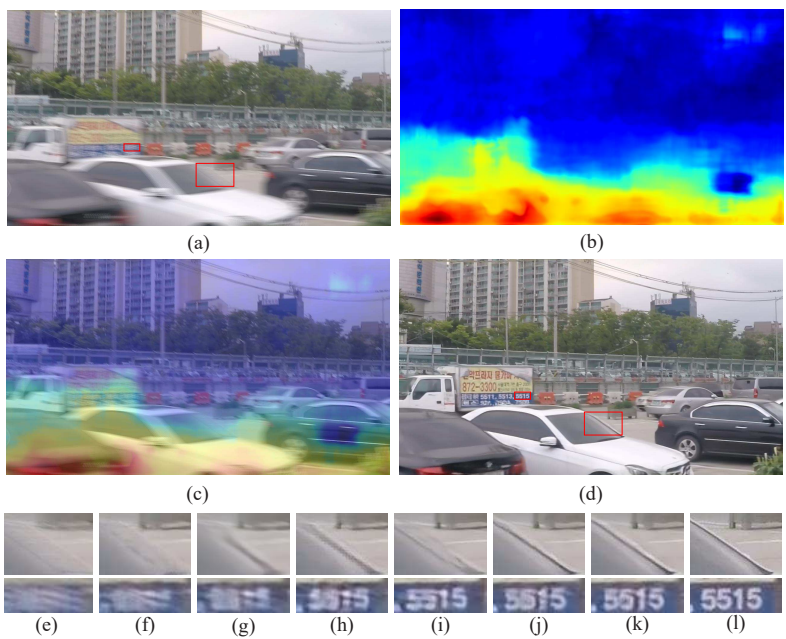}}
	\caption{Visualization results of GoPRO. (a) Blurry frame. (b) the estimated MMP. (c) Overlaying (b) on (a). (d) Result from MMP-RNN. From (e) to (l) are, blurry input, DBN, IFIRNN, ESTRNN, proposed method w/o MMP, MMP-RNN, and the sharp frame, respectively.  }
	\label{fig:gopro}
\end{figure}

\noindent  \textbf{Benchmark Results} 
The results of our method with SoTA lightweight image~\cite{Tao2018} and video deblurring methods on GOPRO is shown in Table~\ref{table:gopro}. We use $A_\# B_\# C_\#F_\#$ to represent $n_a,n_b$,$n_c$ and the length of the input image sequence in our model. We evaluate the image quality in terms of PSNR \cite{Hore2010} and SSIM. We also measure the computational cost for each model when processing one 720P frame in terms of GMACs. Running time (s) of one 720P frame is also listed. For IFIRNN, $c2$ refers to the dual cell model and $h3$ refers to `three times' of hidden state iteration. Our $A_{7} B_{8} C_{16}F_{8}$ model outperforms the other methods with only 204.19 GMACs. The visual comparison of the results are shown in Fig.~\ref{fig:gopro}. 

\begin{figure}[t!]
	\centering
	{\includegraphics[width=0.85\columnwidth]{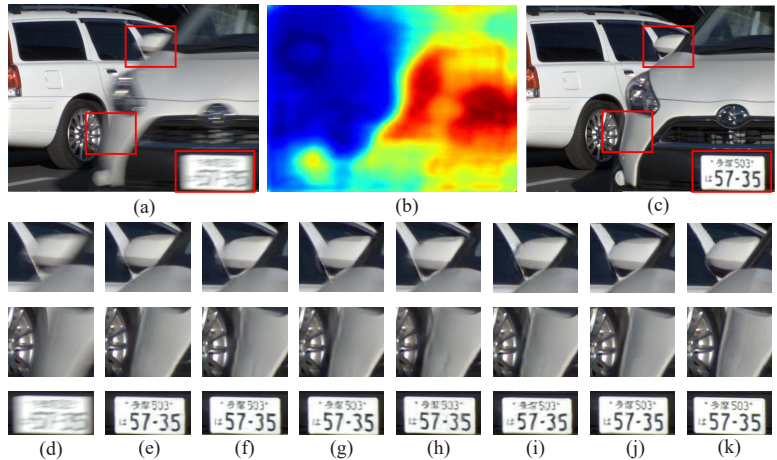}}
	\caption{Visualization results of BSD. a) Blurry frame. (b) the estimated MMP. (c) Result from MMP-RNN. From (d) to (k) are, blurry input, DBN, IFIRNN, ESTRNN, proposed method w/o MMP, MMP-RNN, and the sharp frame, respectively.}
	\label{fig:bsd}
\end{figure}

The results on BSD of the proposed method with other SoTA methods are shown in Table~\ref{table:bsd_results}. Here, we use the $A_{8} B_{9} C_{18}F_{8}$ model. The visualization results of the video deblurring on BSD are shown in Fig.~\ref{fig:bsd}. Our method outperforms SoTA methods with less computational cost.

\setlength{\tabcolsep}{16pt}
\begin{table}[t!]
	\begin{center}
		\caption{Quantitative results on BSD 2ms-16ms.}
		\label{table:bsd_results}
		\scalebox{1.0}{
			\begin{tabular}{lccc}
				\hline
				Model & PSNR & SSIM & GMACs \\
				\hline
				DBN \cite{Su2017}& 31.33 & 0.9132 & 784.75 \\
				IFIRNN ($c2h3$) \cite{Nah2019} & 31.59 & 0.9209 & 217.89 \\
				ESTRNN ($C_{90}B_{10}$) \cite{Zhong2020}& 31.80 & 0.9245 & 215.26\\
				CDVD-TSP \cite{Pan2020} & 32.06 & 0.9268 & 5122.25 \\
				\hline
				\hline
				MMP-RNN ($A_{8} B_{9} C_{18}F_{8}$) &32.79 & 0.9365 & 247.41 \\
				MMP-RNN ($A_{9} B_{10} C_{18}F_{8}$) &\textbf{32.81} & \textbf{0.9369} & 264.52 \\
				\hline
			\end{tabular}
			}
	\end{center}
\end{table}
\setlength{\tabcolsep}{1.4pt}

\subsection{Ablation Study}
\noindent  \textbf{Network Structure} 
We conduct ablation tests on the proposed method with $A_{9} B_{10} C_{18}$ model on GOPRO and $A_{8} B_{9} C_{18}$ model on BSD. We focus on three parts, the MMAM, motion magnitude loss and the transmission of non-deblurred features as shown in Table~\ref{table:ablation}. On GOPRO, the MMAM with MMP can improve PSNR by 0.31~dB. Together with motion magnitude loss, the prior can improve the score by 0.39~dB. If we remove all the components, the PSNR may drop by 0.58, which can indicate the effectiveness of the proposed method. As for BSD, the PSNR significantly increased by 0.63~dB after fusing prior using MMAM. The motion magnitude loss can improve PSNR by 0.09~dB.

\setlength{\tabcolsep}{8pt}
\begin{table}[t!]
	\begin{center}
	\caption{Ablation tests. NDF refers the transmission of non-deblurred features.}\label{table:ablation}
		\begin{tabular}{ccc|cc|cc}
			\hline
			\multicolumn{3}{c|}{Model} & \multicolumn{2}{|c|}{GOPRO} & \multicolumn{2}{c}{BSD}\\
			\hline
			MMAM & $\mathcal{L}_{MM}$ & NDF & PSNR & GMACs & PSNR & GMACs \\
			\hline
			\checkmark & \checkmark & \checkmark & \textbf{32.64} & 264.52 & \textbf{32.79} & 247.41\\
			\ding{53}  & \checkmark & \checkmark & 32.33 & 225.34 & 32.16 & 208.23\\
			\ding{53}  & \ding{53} & \checkmark & 32.25 & 225.34 & 32.07 & 208.23 \\
			\ding{53}  & \ding{53} & \ding{53} & 32.06 & 227.21 & 32.03 & 210.09\\
			\hline
		\end{tabular}
	\end{center}
\end{table}

\setlength{\tabcolsep}{10pt}
\begin{table}[t!]
\begin{center}
\caption{Influence of different types of prior. I. Ground truth, II.Ground truth of flow magnitude of center frame, III. Normalizing each ground truth map to 0$\sim$1, IV. Replacing the MMP and MMAM by spatial-self-attention \cite{CBAM}. V. None, VI. Estimated from MMP-Net.  }
\label{table:priortype}
\scalebox{1.0}{
\begin{tabular}{lcccccc}
				\hline
				Prior type & I & II & III & IV & V & VI \\
				\hline
				PSNR  & \textbf{30.54} & 30.47 & 30.41 & 30.37 & 30.17 & 30.48 \\
				\hline
			\end{tabular}
			}
\end{center}

\end{table}

\noindent  \textbf{Influence of Prior}
We also did ablation tests with different types of priors using $A_{3}B_{4}C_{80}F_{5}$ model on GOPRO. As shown in Table~\ref{table:priortype}, we tried the ground truth MMP, the ground truth of flow magnitude of the center frame (from $FL_{c1}$ and $FL_{cN}$, where $c$ refers to the center frame), the normalized ground truth map ($M$/max($M$)), without MMP and the MMP from MMP-Net. We noticed that the ground truth as an upper boundary can increase the score by 0.37~dB. The MMP from MMP-Net can increase the PSNR by 0.31~dB. On the other hand, using the flow magnitude of the center frame will decrease the PSNR by 0.07~dB. The contour of the center frame magnitude only corresponds to the center frame, the attentive field cannot cover the whole blurry object. Normalizing the MMP by the maximum value of each MMP may lose temporal information, especially for some sharp images, the whole MMP value may be close to 1, which may influence the performance. We also compare the proposed method to spatial-self-attention \cite{CBAM}. Our proposed method use a supervised approach to tell the network where to concentrate and the results outperform vanilla spatial-self-attention.


\section{Conclusion}

In this paper, we proposed a motion magnitude prior for deblurring tasks. We built a dataset of blurry image and motion magnitude prior pairs and used a compact network to learn by regression. We applied the prior to video deblurring task, combining the prior with an efficient RNN. The prior is fused by motion magnitude attentive module and motion magnitude loss. We also transmitted the features before deblurring with features after deblurring between RNN cells to improve efficiency. We tested the proposed method on GOPRO and BSD, which achieved better performance on video deblurring tasks compared to SoTA methods on image quality and computational cost. 

\noindent  \textbf{Acknowledgements} This paper is supported by JSPS KAKENHI Grant Numbers 22H00529 and 20H05951. 

\clearpage
%
%
\bibliographystyle{splncs04}
\bibliography{egbib}
\end{document}